\documentclass[sigconf]{aamas}

\usepackage{balance} 

\doi{WTKR8312}

\makeatletter
\gdef\@copyrightpermission{
  \begin{minipage}{0.2\columnwidth}
   \href{https://creativecommons.org/licenses/by/4.0/}{\includegraphics[width=0.90\textwidth]{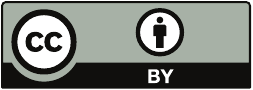}}
  \end{minipage}\hfill
  \begin{minipage}{0.8\columnwidth}
   \href{https://creativecommons.org/licenses/by/4.0/}{This work is licensed under a Creative Commons Attribution International 4.0 License.}
  \end{minipage}
  \vspace{5pt}
}
\makeatother

\setcopyright{ifaamas}
\acmConference[AAMAS '26]{Proc.\@ of the 25th International Conference
on Autonomous Agents and Multiagent Systems (AAMAS 2026)}{May 25 -- 29, 2026}
{Paphos, Cyprus}{C.~Amato, L.~Dennis, V.~Mascardi, J.~Thangarajah (eds.)}
\copyrightyear{2026}
\acmYear{2026}
\acmDOI{}
\acmPrice{}
\acmISBN{}

\acmSubmissionID{1076}

\title{Assessing VLM-Driven Semantic-Affordance Inference for Non-Humanoid Robot Morphologies}

\author{Jess Jones}
\affiliation{
  \institution{Bristol Robotics Laboratory, University of Bristol}
  \city{Bristol}
  \country{United Kingdom}}
\email{jess.jones@bristol.ac.uk}

\author{Sabine Hauert}
\authornote{These authors contributed equally to this work.}
\affiliation{
  \institution{Bristol Robotics Laboratory, University of Bristol}
  \city{Bristol}
  \country{United Kingdom}}
\email{sabine.hauert@bristol.ac.uk}

\author{Raul Santos-Rodriguez}
\authornotemark[1]
\affiliation{
  \institution{University of Bristol}
  \city{Bristol}
  \country{United Kingdom}}
\email{enrsr@bristol.ac.uk}

\begin{abstract}
Vision-language models (VLMs) have demonstrated remarkable capabilities in understanding human-object interactions, but their application to robotic systems with non-humanoid morphologies remains largely unexplored. This work investigates whether VLMs can effectively infer affordances for robots with fundamentally different embodiments than humans, addressing a critical gap in the deployment of these models for diverse robotic applications. We introduce a novel hybrid dataset that combines annotated real-world robotic affordance-object relations with VLM-generated synthetic scenarios, and perform an empirical analysis of VLM performance across multiple object categories and robot morphologies, revealing significant variations in affordance inference capabilities. Our experiments demonstrate that while VLMs show promising generalisation to non-humanoid robot forms, their performance is notably inconsistent across different object domains. Critically, we identify a consistent pattern of low false positive rates but high false negative rates across all morphologies and object categories, indicating that VLMs tend toward conservative affordance predictions. Our analysis reveals that this pattern is particularly pronounced for novel tool use scenarios and unconventional object manipulations, suggesting that effective integration of VLMs in robotic systems requires complementary approaches to mitigate over-conservative behaviour while preserving the inherent safety benefits of low false positive rates.
\end{abstract}


\newcommand{\BibTeX}{\rm B\kern-.05em{\sc i\kern-.025em b}\kern-.08em\TeX}

\begin{document}


\pagestyle{fancy}
\fancyhead{}


\maketitle 


\section{Introduction}

Bringing robotic systems out of controlled laboratory settings into dynamic, real-world environments remains a fundamental challenge on the path toward a future where smart machines are ubiquitous within human society. Achieving robust and adaptive autonomy requires robots to not only perceive their surroundings but also intuitively understand how they can interact with objects and other robots. This is particularly true for future embodied AI, likely heterogeneous in both form and function, with diverse robots developed for specific tasks, often by different commercial entities \cite{deroy2024shared}. Enabling adaptive, collaborative behaviours within such heterogeneous multi-robot systems necessitates a foundational approach to grounding robot interactions.

We hypothesise that grounding these interactions through the concept of affordances \cite{gibson2014theory} can substantially improve their capacity to learn and generalise collaboration strategies. Affordances, defined as the actionable possibilities an environment offers an agent, intrinsically link perception to potential actions. For a robot, understanding affordances means recognising not just what an object is but how it can be interacted with given the robot's unique physical capabilities and the current task. This shifts the paradigm from basic object classification and localisation to dynamic, context-aware interaction, paving the way for more flexible and intelligent robotic systems.

Our broader research proposes to leverage the semantic encoding power of VLMs to enhance the scene-representation of each robot with semantic-affordance descriptions of the objects it encounters, linking its unique capabilities to its environment, and providing a framework from which collaborative interactions can be derived in multi-robot settings in the future. While VLMs show promise, their application to robotics presents a critical challenge, rooted in the fact that their training data is predominantly human-centric \cite{milliere2024anthropocentric}. This raises a fundamental question, which this paper aims to address through empirical analysis: Can VLMs effectively infer affordances for robots with fundamentally different embodiments than humans? We use a novel hybrid dataset, curated for non-humanoid robotic affordance inference, to investigate the performance of VLMs in supporting the zero-shot affordance characterisation of objects across a breadth of diverse contexts, including household items, food products, environmental clean-up, and construction materials. Performance is evaluated on multiple non-humanoid robot morphologies to understand the generalisability and limitations of these models. Our research reveals variations in affordance inference capabilities, identifying biases where VLMs tend toward conservative predictions, characterised by low false positive rates but notably high false negative rates. This conservative bias, while minimising erroneous actions, means robots may underutilise their true capabilities, missing valid interaction and collaboration opportunities, particularly for novel tool use or unconventional manipulations.

Our findings provide critical insights into the practical deployment of VLMs in real-world multi-robot systems. By quantifying the performance variations and identifying the nature of VLM biases in affordance understanding for non-humanoid forms, this work lays the groundwork for developing more robust and reliable embodied AI. The enriched scene-representations, which can be derived from the affordance characterisations of this approach, capture not only what and where objects are, but how the local environment can be interacted with given an arbitrary robot morphology. This will be leveraged in future work to inform action priors that ground learning task decomposition and decision-making strategies in collaborative multi-robot settings, integrating both intrinsic robot abilities and situational demands.

\section{Related Work}
The ability of smart machines to perceive and interact with their environment is fundamental to achieving robust autonomous behavior. A cornerstone concept in this regard is that of affordances \cite{gibson2014theory}. Affordances refer to the actionable possibilities that the environment offers to an individual, which Gibson argues, is not exclusive to humans, but extends to all animals based on a fundamental animal-environment interaction system \cite{gibson2014ecological}. The framing of affordances provides a model which intrinsically links perception and morphology to action. For robotic systems, understanding affordances is crucial for tasks ranging from object manipulation and navigation, to collaborative multi-robot, and human-robot interaction \cite{ardon2020affordances}. Rather than merely identifying objects, robots must infer how they can interact with those objects given their own physical capabilities and the task at hand.

Traditionally, affordance perception in robotics has relied on hand-engineered features, geometric reasoning, or supervised learning on curated datasets specific to particular objects, robots, and tasks \cite{chen2023survey, jamone2016affordances}. While effective in constrained environments, these approaches often struggle with generalisation to novel objects, unseen environments, or diverse robot morphologies. The high cost of data annotation and the limited scalability of these methods present significant barriers to deploying robots in unstructured, dynamic real-world settings.

The recent emergence of large-scale VLMs has impacted various fields within artificial intelligence, including computer vision and natural language processing \cite{ghosh2024exploring}. Trained on vast amounts of internet-scale image-text data, VLMs have shown capabilities in zero-shot generalisation, common-sense reasoning, and cross-modal understanding \cite{radford2021learning, parelli2023clip}. Their ability to connect visual input with high-level semantic concepts, often expressed in natural language, presents a compelling opportunity to address the limitations of traditional affordance perception methods in robotics.

Initial research has begun to explore the application of VLMs to robotic affordance prediction, primarily by leveraging their inherent understanding of human-object interactions \cite{do2018affordancenet, bahl2023affordances, qian2024affordancellm, yuan2024robopoint, li2024one, lee2024affordance}. These works often frame affordance as a segmentation, localisation or trajectory modelling problem, identifying regions of objects that afford certain human actions (e.g. "grasp," "cut," "pour"). However, a critical gap remains in understanding how these human-centric affordance concepts translate to non-humanoid robotic embodiments. Robots possess diverse manipulators, grippers, and end-effectors, with physical capabilities that may differ significantly from human hands. This disparity raises fundamental questions about the direct applicability and generalisation capabilities of VLMs across the broader spectrum of existing and future robotic morphologies, and their unique interaction possibilities. This work represents an initial exploration of this unexplored territory, investigating the effectiveness of VLMs in inferring affordances for robots with fundamentally different embodiments than humans, and through empirical analysis, understanding the systematic biases that may arise from their human-centric training data.

Building upon advancements in foundation models and their applications, a body of work has focused on the creation and utilization of rich semantic scene representations. Rather than simply identifying individual objects or their spatial location, these representations aim to capture a more holistic understanding of the environment, including the relationships between objects, their attributes, and even the higher-level context of a scene. Recent work in this area, leveraging open-vocabulary concepts \cite{chen2023open,kassab2024language,fu2024scene} and multimodal inputs \cite{chang2023goatthing, ahn2024autort,gu2023rt}, has significantly enhanced robots' abilities to navigate and plan in novel environments. Our research explores introducing an additional dimension to scene representation by investigating how different robotic agents can meaningfully interact with those environments, specifically by inferring actionable affordances using VLMs. This bridges the gap between static scene understanding and dynamic, embodied interaction.

\section{Affordance Inference Pipeline}
\begin{figure*}[t]
\centering
\includegraphics[width=0.95\textwidth]{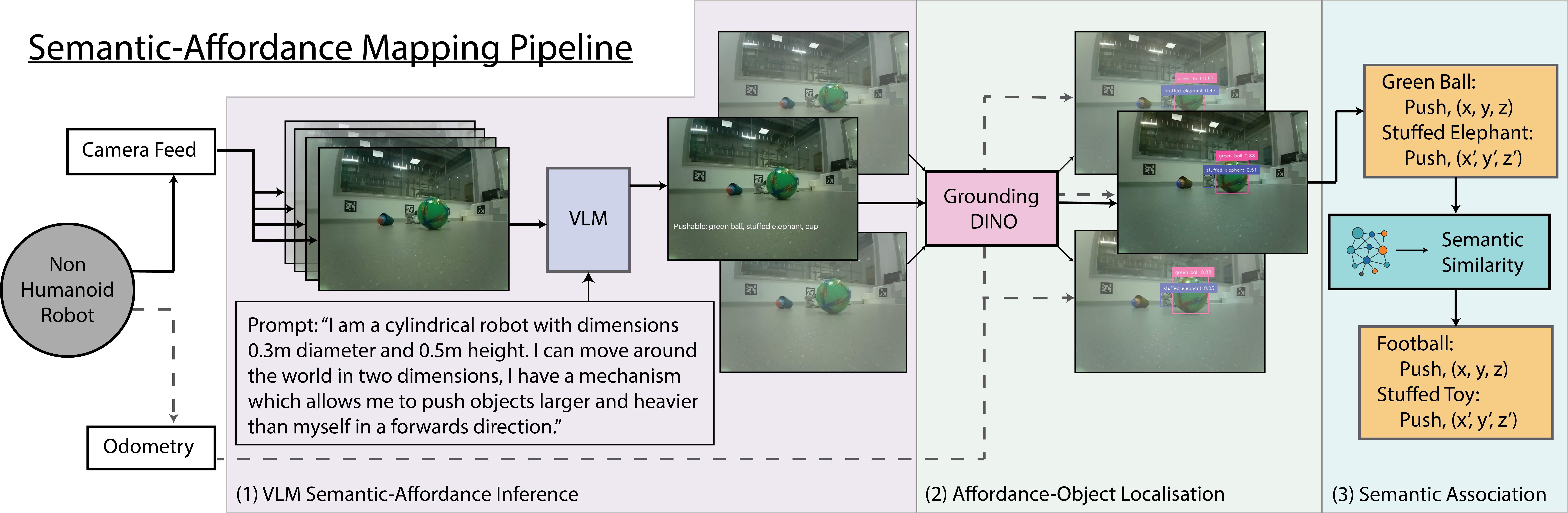}
\caption{An illustration of our Semantic-Affordance Mapping pipeline applied to a non-humanoid robot (1) The camera feed of the robot is sampled every $n$ frames and provided as input to the VLM alongside a natural language description of the physical properties of the robot. The VLM is tasked to identify objects in the frame and the affordances available to the robot, given its physical description. A structured output of affordance-object relations is returned. (2) Identified objects are passed to GroundingDINO with frame $n$ and two frames either side ($n-k$, $n+k$) to establish bounding boxes in the image plane. These bounding boxes are used with the recorded robot odometry to triangulate object positions in world space, and form the semantic-affordance mapping tuple: $\langle$Object, Affordance, Location$\rangle$. (3) Semantic similarity measures are performed against existing objects in the scene graph within distance $d$ to consolidate variance in the VLM object labels.}
\label{fig1}
\end{figure*}

This section details our experimental pipeline and demonstrates a potential application for VLM-driven affordance inference within multi-robot systems.

\paragraph{VLM Semantic-Affordance Inference}
We begin by defining natural language prompts for the VLM with a description of the robot's physical properties. This enables the VLM to ground its understanding of affordances within the specific context of the robot's embodiment. We are interested in exploring the generalisable inference capabilities of these models, and therefore restrict these prompts to concise descriptions which could be applicable to any task domain. For instance, in our experiments, we provide descriptions such as:

"I am a cylindrical robot with dimensions 0.3m diameter and 0.5m height. I can move around the world in two dimensions. I have a mechanism which allows me to scoop objects from the ground if the object weighs less than 1 kilogram and is no wider than 0.3m."

For our experiments with non-humanoid robot morphologies, we consider a heterogenous set of six robots with distinct capabilities: Scoop, Push, Lift, Cut, Pick, and Collect. With a view towards future works where this pipeline will be deployed on real-world multi-robot collaboration problems, we take the DOTS \cite{jones2022dots} robots as a base, and envisage modifications to facilitate these capabilities. A complete set of robot prompts is provided in the associated datasets linked in Section \ref{sec:code}.

As the robot navigates its environment, the on-board imaging system captures visual data. Frames are sampled at rate $n$ to balance computational load with the need for fresh environmental information. We choose $n=24$ to capture fresh information every second. Each sampled image along with the robot's specific physical description is then transmitted to the VLM via an API call. The VLM outputs explicitly identify detected objects within the sampled image and provide corresponding affordance characterisations relevant to the robot's described capabilities. Outputs are formatted as dictionaries of individual affordances, and a corresponding list of identified objects which afford this capability.

\paragraph{Localisation}
To anchor the VLM's 2D image-based object-affordance inferences into a unified, global 3D reference frame, we employ a multi-view localisation strategy. For each VLM-identified object and its associated affordances, we leverage nearby frames captured by the on-board camera, specifically, given a detection event in frame $n_i$ we sample frames $n_i \pm k$. We set $k=0.5*$frame-rate. The semantic object descriptions generated by the VLM (e.g. "cup," "plastic pipe") are passed to GroundingDINO \cite{liu2024grounding}, a zero-shot object detection model, to generate 2D bounding boxes for these objects across the sequence of frames. By combining these 2D detections with the robot's odometry and camera calibration parameters, we triangulate the global 3D position of each object.

\paragraph{Semantic-Association}
The inherent stochastic nature of generative model outputs can lead to inconsistencies where the same physical object might be identified with slightly varying labels (e.g. "green ball" vs. "football") or conflicting affordance predictions across different observations (e.g. "grasp", "grip-able"). To mitigate this and ensure the construction of a robust, consistent overall scene representation, we implement a semantic-association mechanism. This mechanism continuously evaluates the semantic similarity of newly detected objects and their affordances against previously labelled entities within a specified spatial proximity $d$ of the current detection's global position. In our experiments, we assign $d <$  L2 norm threshold of $0.1$.

Specifically, when a new object-affordance pair is localised, sentence transformer embeddings are generated for both object and affordance labels, and compared via cosine similarity to those of existing objects in the scene graph that fall within distance $d$. If semantic similarity exceeds a predefined threshold ($\tau = 0.45$) and the global positions of the objects are sufficiently close, the new detection is associated with the existing object. This fusion process resolves conflicting labels by, for instance, merging "cup" and "mug" detections into a single entity or by consolidating "pickable" and "graspable" affordances for the same object. This ensures that the scene representation is coherent, reduces noise from transient VLM outputs, and facilitates long-term mapping and consistent interaction planning.

\begin{table}[t]
    \begin{tabular}{|c|c|c|c|}
        \hline
        \textbf{Source} & \textbf{Unique Objects} & \textbf{Instances} \\
        \hline
        Real & 14 & 246 \\
        \hline
        Synthetic & 86 & 528 \\
        \hline
        Total & 100 & 774 \\
        \hline
    \end{tabular}
    \caption{Composition of our novel dataset. The table presents a breakdown of the 774 labelled object instances, categorized by their real and synthetic origins. This combined dataset forms the basis for all experimental evaluations.}
    \label{tab:dataset}
\end{table}

\begin{table}[t]
    \begin{tabular}{|c|c|c|}
        \hline
        \multicolumn{3}{|c|}{Humanoid Robot} \\
        \hline
        \textbf{Affordance} & \textbf{Unique Objects} & \textbf{Instances}  \\ 
        \hline
        Lift & 93 & 736 \\
        \hline
        Pick & 92 & 735 \\
        \hline
        Push & 93 & 736 \\
        \hline\hline
        \multicolumn{3}{|c|}{Non-Humanoid Robots}\\
        \hline
        \textbf{Affordance} & \textbf{Unique Objects} & \textbf{Instances}  \\ 
        \hline
        Scoop & 76 & 499\\
        \hline
        Lift & 1 & 11\\
        \hline
        Collect & 76 & 507\\
        \hline
        Push & 92 & 735\\ 
        \hline
        Pick & 78 & 533\\
        \hline
        Cut & 39 & 253\\
        \hline    
    \end{tabular}
    \caption{Distribution of affordance-object pairs in the dataset. The table itemizes both unique pairs and total instances, which sum to 4,745 instances overall. For analytical clarity, data for the baseline humanoid robot is presented separately from the six non-humanoid robots.}
    \label{tab:aff_dataset}
\end{table}

\section{Evaluation Methodology}
With this work, we provide a preliminary novel hybrid dataset specifically designed for affordance characterisation in the context of non-humanoid robot morphologies. The dataset is presented not as a comprehensive collection but as an initial contribution to the study of affordance inference for non-humanoid robots. This collection comprises both synthetic and real-world video sequences, captured from the on-board perspectives of non-humanoid robots in environments consisting of everyday objects relevant to diverse real-world scenarios. The dataset is curated to reflect the challenging environments we explore: household settings, construction sites, logistics warehouses, and environmental clean-up. One hundred unique objects have been selected for diversity and include items such as: Book, Metal Rod, Table Tennis Bat, Stuffed Toy, Tissue, and Pebble. A summary of the unique object counts, and labelled object instances within the dataset is presented in Table \ref{tab:dataset}, and an overview of the affordance-object pairs in the dataset is presented in Table \ref{tab:aff_dataset}. For both synthetic and real-world datasets, affordance ground truth is established through human annotation. Links to the complete dataset are provided in section \ref{sec:code}.

\paragraph{Synthetic Data} Our synthetic material comprises a collection of videos generated using Google's Veo3 model. These videos feature a wide range of objects across a diverse set of environments from the on-board perspective of a robot with non-humanoid morphology. We also include a collection of isolated synthetic images of individual objects from these contexts, designed for direct VLM analysis of affordance capabilities independent of complex scene clutter. All synthetic materials are manually labelled to mitigate some of the risks of inherent bias, and we observe a difference of $\pm 0.05$ in mean object-affordance inference scores between the real and synthetic datasets, indicating that any bias introduced by the generative models is negligible in the scope of this work.

\paragraph{Real-World Data} To assess the practical applicability and generalisation of our findings, we provide real-world videos captured using the DOTS robots. The material is captured in a controlled but cluttered indoor arena environment, designed to mimic elements of the target scenarios. 

Our evaluation across these datasets demonstrates the intrinsic capabilities of the VLM-driven affordance characterisation pipeline across a broad spectrum of controlled conditions and diverse robot morphologies. The real-world materials serve to validate that these capabilities translate effectively to tangible robotic systems within a physical environment, and with limited imaging and processing capabilities. 

We consider three VLMs in our assessment using the providers' default hyper-parameters and conduct five independent trials to identify standard deviation in the results. The models selected are noted by their respective API keys, and represent the latest image-capable public releases as of September 2025: 
\begin{itemize}
    \item GPT: gpt-5
    \item Gemini: gemini-2.5-pro
    \item Claude: claude-opus-4-1-20250805
\end{itemize}

We examine affordances for six independent non-humanoid robot morphologies, each with unique action capabilities: Scoop, Push, Pick, Lift, Cut, Collect. 
The VLMs are prompted with a simple description of a given robot and its unique physiological properties, for example, the 'Lifting' robot: 

"I am a cylindrical robot with dimensions 0.3m diameter and 0.5m height. I can move around the world in two dimensions. I have a mechanism which allows me to lift objects vertically upwards as long as I can get underneath them."

In addition, we consider a humanoid robot as a baseline for comparison. To provide a direct comparison for this baseline, we constrain our evaluation labels for the humanoid to three capabilities (Push, Pick, Lift) which reflect the capabilities of three of our six non-humanoid robots. We provide a minimal prompt to the VLMs describing this robot: 

"I am a humanoid robot, I have two arms and two legs, and have the same capabilities as a human."

All experiments were conducted offline on an Apple M3 Pro with 18GB memory. A full package list with versions is provided with the code repository linked in Section \ref{sec:code}. The DOTS robot \cite{jones2022dots} hardware was used for the real data collection.

\begin{figure*}[t]
\centering
\includegraphics[width=1\textwidth]{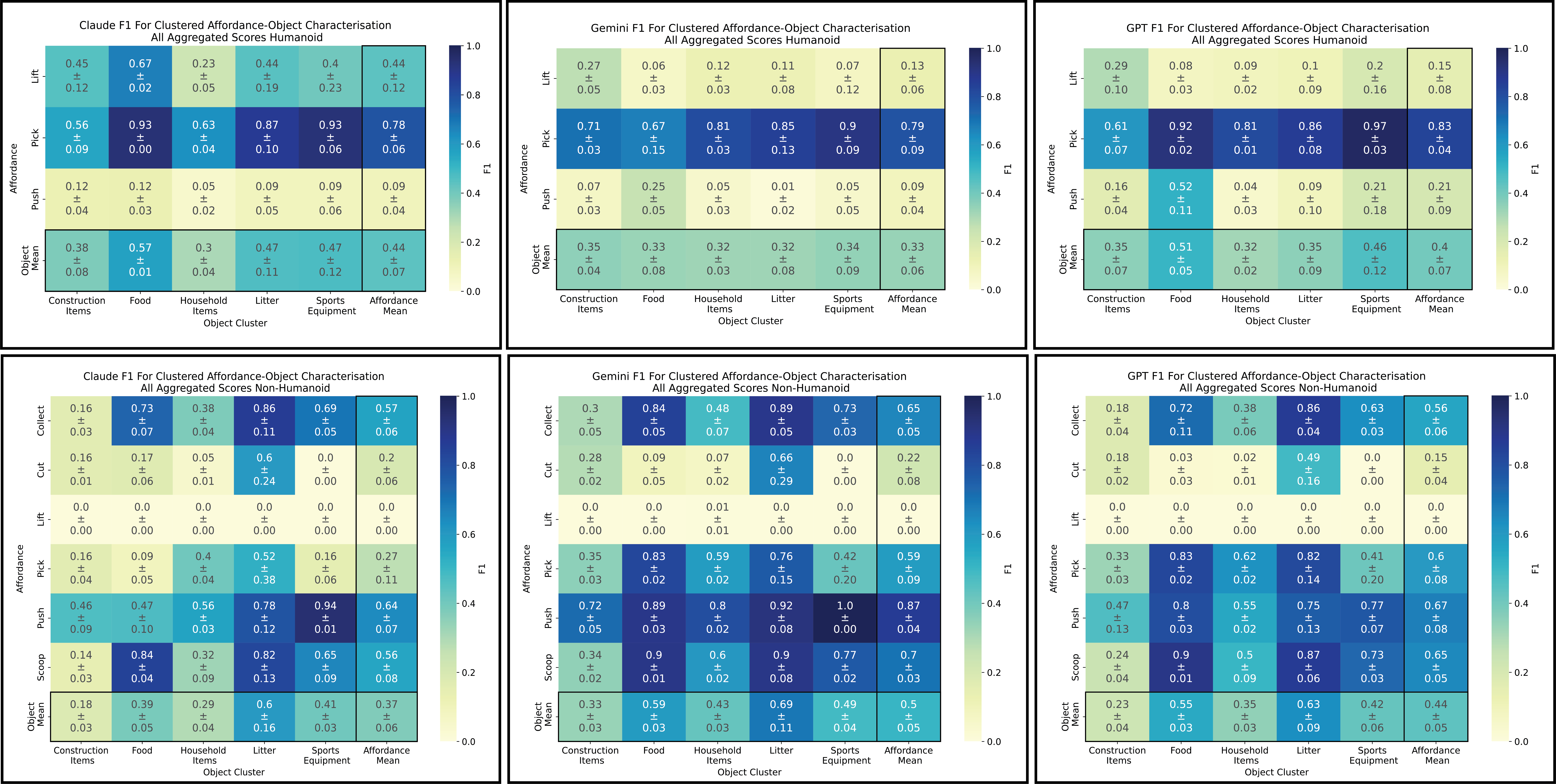}
\caption{Affordance-object inference F1 scores and standard deviation over five independent trials. Objects have been clustered into five distinct groups. The top row describes the results produced by the baseline humanoid robot, and the bottom row the non-humanoid robots. Of the three affordances we examine, the humanoid baseline captures the 'Pick' affordance, synonymous with 'Grasp', but tends to default to out-of-scope human-like affordances in place of 'Lift' in the case of Gemini and GPT and 'Push' across all models, leading to a high-false negative rate and lower overall performance. Improvements are observed in non-humanoid robots for the 'Push' affordance, indicating that constrained descriptions of the robots capabilities are beneficial for novel tool use. Notably, Claude's inference capabilities of the 'Pick' affordance dramatically diminishes for non-humanoid robots, suggesting a lesser capacity for embodiment generalisation. The non-humanoid robot which affords 'Lift' underperforms due to a criterion of objects being above the robot chassis, exposing the weaknesses in spatial awareness and generalised context awareness.}
\label{fig2}
\end{figure*}

\subsection{Metrics}
Our evaluation methodology assesses the performance of the VLMs in accurately inferring affordance labels and corresponding objects. For each sampled frame $f \in \mathcal{F}$, where $\mathcal{F}$ is the set of all evaluated frames, VLM annotations are compared against human-annotated ground truth labels. 

\paragraph{Annotation Representation} Let an individual annotation, whether ground truth or VLM-generated, be represented as a tuple $(a, O)$, where:
\begin{itemize}
    \item $a$ is one of the affordance labels associated with a given robot morphology.
    \item $O = \{o_1, o_2, \ldots, o_m\}$ is a set of $m$ object labels associated with the affordance $a$.
\end{itemize}
For a given frame $f$, let $\mathcal{GT}_f = \{(a_{GT,i}, O_{GT,i})\}_{i=1}^{N_{GT}}$ be the set of $N_{GT}$ ground truth annotations representing unique affordances and corresponding object relations, and $\mathcal{VLM}_f = \{(a_{VLM,j}, O_{VLM,j})\}_{j=1}^{N_{VLM}}$ be the set of $N_{VLM}$ VLM-predicted annotations. 

\paragraph{Semantic Similarity and Pairwise Validation} For each sampled frame $f \in \mathcal{F}$:
Let $\text{sim}(l_1, l_2)$ be the semantic similarity function between two labels $l_1$ and $l_2$ (e.g. cosine similarity of word embeddings), yielding a value in $[0, 1]$. We use the SBERT `all-MiniLM-L6-v2` sentence transformer to generate our word embeddings, and by sampling a selection of typical VLM affordance assignments we find a semantic similarity threshold of $\tau = 0.45$ strikes a suitable balance between precision and recall.

\paragraph{Evaluation Process} Evaluation for each VLM prompt response corresponds to a specific robot morphology within the frame.
\begin{itemize}
    \item Let $\mathcal{A}_{GT}^{(f,r)}$ be the list of ground truth affordance labels relevant to a specific robot morphology $r$ in frame $f$.
    \item Let $\mathcal{O}_{GT}^{(f, a)}$ be the list of ground-truth object labels relevant to each affordance $a \in \mathcal{A}_{GT}^{(f,r)}$.
    \item Let $\mathcal{A}_{VLM}^{(f, r)}$ be the list of affordance labels predicted by the VLM, relevant to a specific robot morphology $r$ in frame $f$.
    \item Let $\mathcal{O}_{VLM}^{(f, a)}$ be the list of object labels associated by the VLM, relevant to each affordance $a \in \mathcal{A}_{VLM}^{(f,r)}$.
\end{itemize}

A triangular similarity matrix $S_a^{(f)}$ is computed, where each element $S_a^{(f)}[i, j]$ represents the semantic similarity between the $i$-th ground truth affordance in $\mathcal{A}_{GT}^{(f,r)}$ and the $j$-th VLM-predicted affordance in $\mathcal{A}_{VLM}^{(f, r)}$:
$S_a^{(f, r)}[i, j] = \text{sim}(\mathcal{A}_{GT}^{(f,r)}[i], \mathcal{A}_{VLM}^{(f, r)}[j])$.

\begin{figure*}[t]
\centering
\includegraphics[width=1\textwidth]{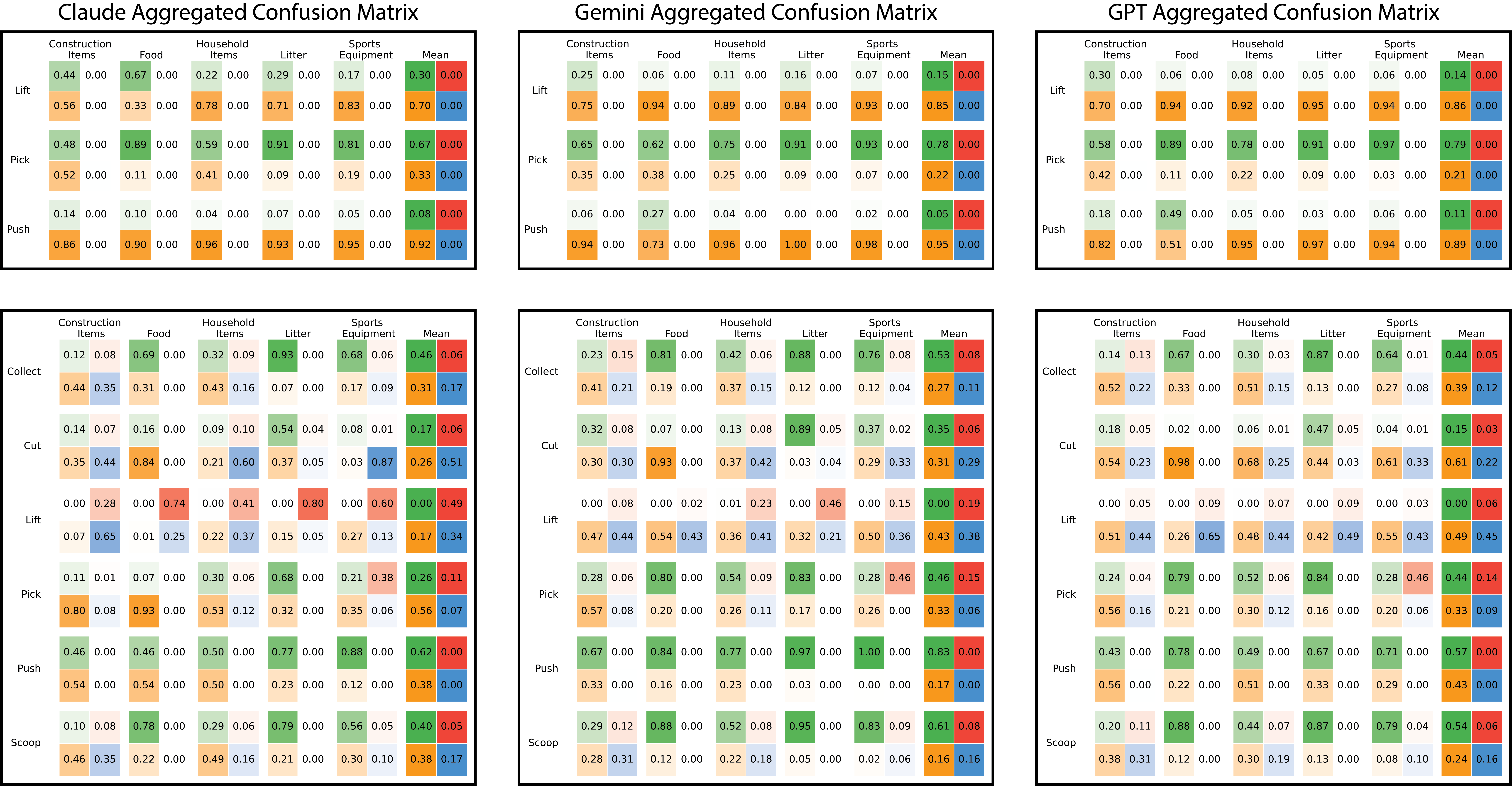}
\caption{Confusion matrices across the three VLMs for aggregated performance of True-Positive (green), False-Positive (red), True-Negative (blue), False-Negative (orange) across the clustered object classes. Opacity in columns (excluding the mean) reflect the weight of the respective score. As with Figure \ref{fig2}, the humanoid robot baseline is presented on the top row, and the non-humanoid robots on the bottom. Note that results are predominantly weighted towards true-positive (correct affordance characterisation), or false negative (conservative bias) for most affordance-object relations with notable outliers being the 'Lift' and 'Cut' affordances for the non-humanoid robots.}
\label{fig3}
\end{figure*}

A set of matched affordance pairs $\mathcal{M}_a^{(f, r)}$ is identified. Each pair $(i, j) \in \mathcal{M}_a^{(f, r)}$ indicates that the $i$-th ground truth affordance in $\mathcal{A}_{GT}^{(f,r)}$ has a semantically similar counterpart in the $j$-th VLM-predicted affordance in $\mathcal{A}_{VLM}^{(f, r)}$:
$\mathcal{M}_a^{(f, r)} = \{ (i, j) \mid S_a^{(f, r)}[i, j] > \tau \}$
For each such match, the corresponding VLM-predicted affordance is denoted as $\hat{a}_{VLM}^{(f)}$.

Object matching is then computed, conditional on affordance matching: For each matched affordance $\hat{a}_{VLM}^{(f)}$:
A triangular similarity matrix $S_o^{(f,\hat{a})}$ is computed. This matrix compares the ground truth objects associated with this affordance $\mathcal{O}_{GT}^{(f, \hat{a})}$ against the objects predicted by the VLM. Each element $S_o^{(f,\hat{a})}[k, l]$ represents the semantic similarity between the $k$-th ground truth object in $\mathcal{O}_{GT}^{(f, \hat{a})}$ and the $l$-th VLM-predicted object in $\mathcal{O}_{VLM}^{(f, \hat{a})}$:
$S_o^{(f,\hat{a})}[k, l] = \text{sim}(\mathcal{O}_{GT}^{(f, \hat{a})}[k], \mathcal{O}_{VLM}^{(f, \hat{a})}[l])$. 

We are interested in understanding the performance of the VLMs in correctly identifying when an object does and does not afford the capabilities of a given robot morphology. In our analysis we consider the full confusion matrix of True Positives (TP), True Negatives (TN), False Positives (FP), and False Negatives (FN) across all frames $f \in \mathcal{F}$. Understanding the balance of precision and recall is important to this assessment, therefore we select the F1 score as our evaluation metric, computed as: 
\[ F_1 = \frac{2 \cdot \text{TP}}{2 \cdot \text{TP} + \text{FP} + \text{FN}} \]

\section{Results and Discussion}
\begin{figure*}[t]
\centering
\includegraphics[width=0.95\textwidth]{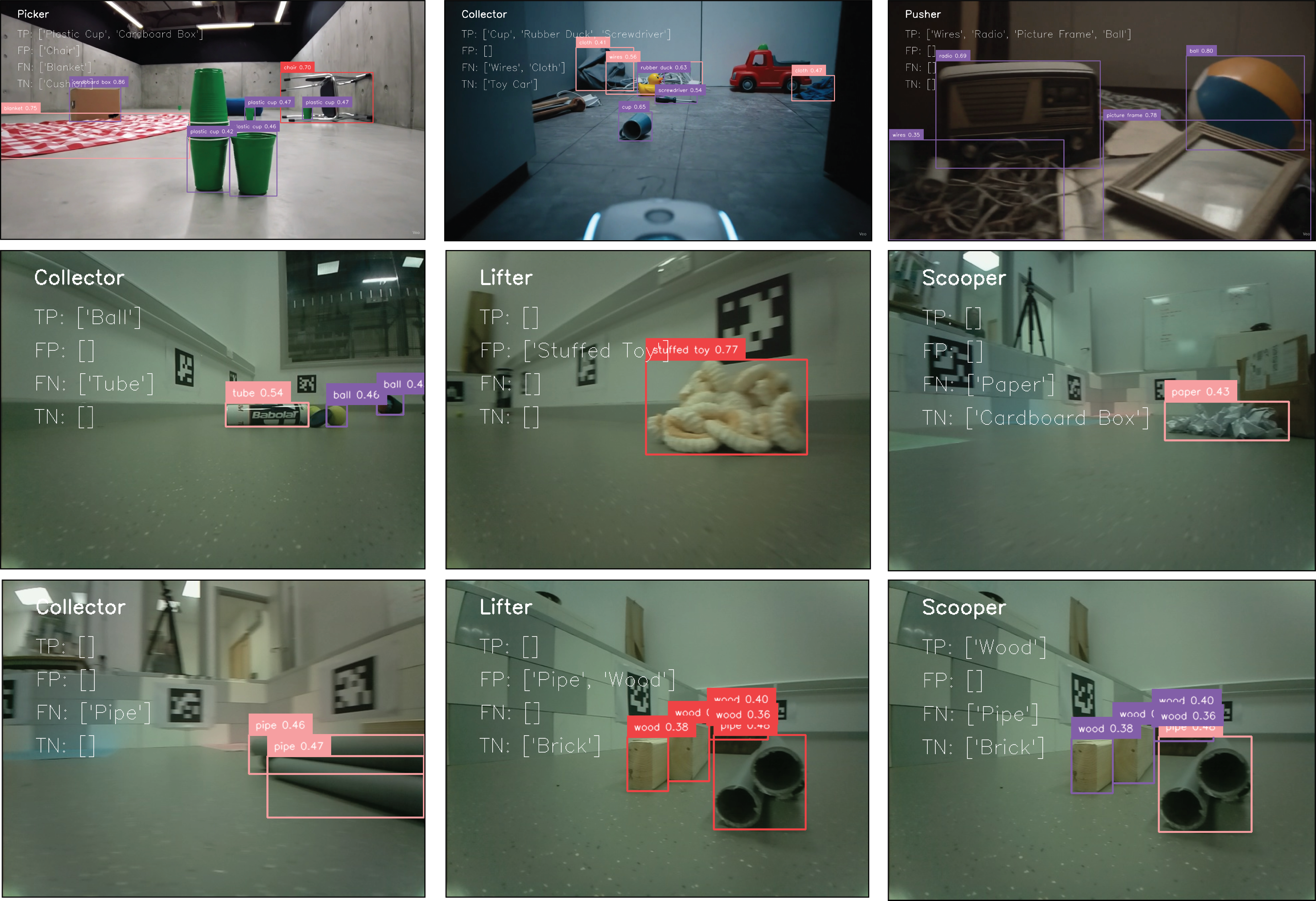}
\caption{Examples of VLM semantic-affordance inference mapped to bounding boxes with GroundingDINO. The top row illustrates examples from the synthetic video dataset, and the middle and bottom rows from the real world data; household items, and construction materials respectively. Affordance and performance classifications have been overlayed to provide evaluation insights in these cases. True-positives are bounded by purple boxes, false-positives by red boxes, and true-negatives by pink boxes.}
\label{fig4}
\end{figure*}

In this section, we analyse the affordance inference performance of the three selected VLMs. Our analysis covers the baseline humanoid robot and the six non-humanoid robot morphologies. Aggregated F1 scores and detailed confusion matrices are presented in Figure \ref{fig2} and Figure \ref{fig3}, respectively, with data clustered into five distinct object categories for clarity. Individual affordance-object scores are available in the data linked in Section \ref{sec:code}. In Figure \ref{fig4} we present frames from the synthetic and real world video data to visualise the performance classifications in our experiments. For the humanoid baseline, the models achieved moderate overall performance, with Claude $(0.53)$ and GPT $(0.51)$ showing better mean affordance-object scores vs. Gemini $(0.36)$. There was a clear performance disparity across the affordances, where all models were generally successful at inferring the 'Pick' action but struggled significantly with 'Push'. For the non-humanoid robots, affordance characterisation was more varied, with Gemini showing the strongest overall results with a mean F1 score of $0.5$. The non-humanoid results highlight a clear trend where affordances explicitly linked to the robot's physical form, such as 'Scoop' and 'Push', were identified far more successfully than actions like 'Cut' and 'Lift', which were defined with more complex spatial and material constraints.

\paragraph{Humanoid Baseline}
The VLMs showed the strongest performance when inferring the 'Pick' affordance, which is synonymous with the common human action of grasping. This was particularly evident for object classes like 'Food' and 'Sports Equipment', which are frequently associated with human interaction in the models training data.

However, an important observation was made across all affordances. Instead of identifying the labelled capabilities, the models frequently defaulted to generating a wide range of plausible, but out-of-scope, human-like actions such as: 
\begin{itemize}
\item {"Throw", ["Tennis Ball", "Rugby Ball"]}
\item {"Squeeze", ["Gray Stuffed Animal"]}
\end{itemize}
Detailed example response logs are provided in the data linked in Section \ref{sec:code}. This tendency to default to general human-centric world knowledge resulted in a high rate of false negatives across all models for the examined affordances (Figure \ref{fig3}, top row). We suggest this reveals an inherent bias: the models will omit a correct, task-relevant affordance if a more common but out-of-scope action seems applicable from their training. This observation requires further investigation and is discussed in more detail in Section \ref{future}.

\paragraph{Non-Humanoid Robots}
In contrast, the non-humanoid robots showed a marked improvement on the equivalent 'Push' affordance (Figure \ref{fig2}, bottom row). This affordance is clearly inferred from the physical description of the robot: "\textit{I have a mechanism which allows me to push objects larger and heavier than myself in a forwards direction.}", but crucially, the models generated significantly fewer out-of-scope affordances. The case is similar for robots with tools to 'Scoop' and 'Collect'. This indicates that providing a concise physical description of the robot successfully constrains inference, grounding its reasoning in the specific capabilities of the robot. This grounding helps overcome the tendency to revert to general human actions, leading to more accurate and relevant affordance characterization. 

However, there is also notable trend of false-negatives across the non-humanoid robots (Figure \ref{fig3}, bottom row). Given the lack of out-of-scope affordance characterisations, this indicates that models tend towards a conservative bias, where they are more likely to assign no affordance than an incorrect affordance. This tendency preserves a low false positive rate, which is beneficial for safety, but raises an important consideration for the effective application of VLMs in scenarios with diverse robot morphologies and unconventional applications.

Despite the benefits of physical grounding, our results also highlight key limitations in the VLMs' reasoning, particularly when faced with unconventional scenarios and novel tool use. 

A significant weakness in spatial reasoning was revealed by the models attribution of the 'Lift' affordance. Even though this action was explicitly constrained to objects above the non-humanoid robot's chassis, Claude and GPT incorrectly identified items on the ground as 'liftable' leading to high rates of false-positives (Figure \ref{fig3}, bottom row). This inability to apply a general concept to a novel physical context demonstrates a fundamental lack of spatial and contextual understanding. Interestingly, this is less evident in the Gemini results, where the model either generated a false-negative (conservative bias) or a true-negative (correct interpretation).

Inspection of the individual affordance-object scores revealed that models also showed weaknesses in constraining inference when considering the 'Cut' affordance. This robot description constrained the valid cut-able materials to paper, plastic or wood. However, there was a tendency across the board to assign cut to other materials types, including 'Golf Ball', 'Screwdriver', 'Paint Pot'. This is perhaps indicative of a lack of materials understanding, and warrants further enquiry.  

We also observe notable performance inconsistencies across different object classes (Figure \ref{fig2}, bottom row). F1 scores were generally higher for 'Food', 'Litter', and 'Sports Equipment', but considerably lower for categories like 'Construction Items'. This further supports the hypothesis that this is a result of the models human-centric training data, where the vast corpus of images depicting humans interacting with food and sports equipment allows for better generalization. Conversely, the models limited exposure to robotic manipulation in domains like construction hinders their ability to infer affordances for less common object interactions, revealing a clear boundary in their inference capabilities.

\section{Conclusion} In summary, for a humanoid robot, the VLMs performed well on common actions like 'Pick'. However, they frequently suggest plausible human actions (e.g. "Throw" a ball). This resulted in a high rate of false negatives, where the correct action was overlooked in favour of a more common, but out-of-scope, human one. This insight highlights the value of considering alternative evaluation techniques, where VLM outputs are validated by a human annotator, with the caveat that consideration must be given to the significant human resourcing required to effectively do this.

In contrast, for non-humanoid robots, providing a clear physical description (e.g. "... I have a mechanism to push objects ...") successfully grounded the models' reasoning, leading to fewer out-of-scope suggestions. This improvement came with a trade-off: the models adopted a conservative bias, often returning no affordances rather than inferring an incorrect one, which also led to false negatives.
The research also highlighted several critical weaknesses in the underlying inference capabilities of the VLMs:

\paragraph{Poor Spatial Reasoning} Models incorrectly identified objects on the ground as 'liftable' for a robot that could only lift objects above itself, demonstrating a failure to apply spatial constraints.

\paragraph{Lack of Material Understanding} 
Models suggested that a robot could 'Cut' inappropriate materials like a golf ball or screwdriver, ignoring specified materials constraints.

\paragraph{Training Data Bias} Performance was consistently better with familiar object categories like 'Food' and 'Sports Equipment' and worse with 'Construction Items', indicating that the models' abilities are limited by their human-focused training data.

\section{Future Works}
\label{future}
Our work proposes using VLMs to support semantic-affordance inference and enhance robot scene representation. In future work these representations can be used to ground multi-robot collaboration, enabling teams of diverse robots to collaborate to solve problems by reasoning about their shared environment and their respective contribution capacity. While we have shown the potential of this approach, our findings also highlight key limitations of using VLMs for affordance inference, presenting clear opportunities for additional streams of research.

\paragraph{Task Constraints} We have observed that VLMs often suggest actions that are plausible in a general sense but irrelevant to the task at hand. Since affordances are inherently task-dependent, a crucial next step is to explore task-conditioned prompting. By supplementing the robot's physical description with a high-level task description (e.g. "tidy the room"), we can investigate whether VLMs can better constrain their inference to output only the most relevant affordances. A preliminary probe, involving a very simple task description on our real data set, shows an immediate uplift in average object-affordance mean of between $0.03 - 0.1$  for the non-humanoid robots, indicating scope to expand this line of enquiry further.

\paragraph{Physical Embodiments} Development of the heterogenous capabilities as an extension to the base DOTS platform is noted as important future work for forthcoming research and real-world execution prototyping.

\paragraph{Validation Approach} We also note that there is value in exploring alternative methods to evaluate the outputs of generative AI models. Instead of providing a set of 'gold-standard' label outputs, which we want the model to generate, we suggest each generated output is validated by a human for the correctness of the affordance characterisation. This would provide insights to understand the extent of noise in the labels and the potential for more complex, 'unexpected' inference capabilities. Naturally, annotation resources and consistency are a potentially limiting factor of this evaluation approach, as significant human validation efforts would be required.  

\paragraph{Inference Speed} A significant bottleneck for real-world deployment is slow inference speed, with API calls in our experiments taking three to ten seconds, prohibiting real-time updates. Future work should focus on accelerating inference, with promising avenues including local, fine-tuned models and model distillation.

\paragraph{Dataset Expansion}
Finally, our study relied on a dataset created specifically for this investigation, which could be further improved and expanded. To accelerate progress, the field requires a large-scale, public benchmark dataset for non-humanoid robot affordance inference. A well-annotated community benchmark would enable rigorous, standardised evaluation of different models and propel the development of more generalizable and robust systems.

\section{Code \& Data Repositories} 
\label{sec:code}
The code and data repository links are made available here: \url{https://bitbucket.org/hauertlab/vlm_driven_non_human_sem_aff}

\begin{acks}
JJ is support by FARSCOPE EPSRC CDT. RSR is funded by the UKRI Turing AI Fellowship [grant number EP/V024817/1]. SH is supported by an EPSRC Open Plus Fellowship.
\end{acks}



\bibliographystyle{ACM-Reference-Format} 
\balance
\bibliography{sample}


\end{document}